\def\year{2022}\relax
\documentclass[letterpaper]{article} 
\usepackage{aaai22}  
\usepackage{times}  
\usepackage{helvet}  
\usepackage{courier}  
\usepackage[hyphens]{url}  
\usepackage{graphicx} 
\usepackage{natbib}  
\usepackage{caption} 
\DeclareCaptionStyle{ruled}{labelfont=normalfont,labelsep=colon,strut=off} 
\usepackage{booktabs}
\usepackage{multirow}
\usepackage{graphicx}
\usepackage{diagbox}
\usepackage{microtype}
\usepackage{amsmath}
\usepackage{amssymb}
\usepackage[switch]{lineno}
\usepackage{algorithm}
\usepackage{algorithmic}

\usepackage{newfloat}
\usepackage{listings}
\floatstyle{ruled}
\newfloat{listing}{tb}{lst}{}
\floatname{listing}{Listing}
\title{Reminding the Incremental Language Model via Data-Free Self-Distillation}
\author{
    
    Han Wang,\textsuperscript{\rm 1, 2}
    Ruiliu Fu,\textsuperscript{\rm 1, 2}
    Chengzhang Li,\textsuperscript{\rm 1, 2}
    Xuejun Zhang,\textsuperscript{\rm 1, 2}
    Jun Zhou,\textsuperscript{\rm 1, 2}
    Yonghong Yan\textsuperscript{\rm 1, 2}
    
}
\affiliations{
    
    \textsuperscript{\rm 1} Key Laboratory of Speech Acoustics and Content Understanding, Institute of Acoustics, China\\
    \textsuperscript{\rm 2} University of Chinese Academy of Sciences, Beijing, China\\
    \{wanghan, furuiliu, lichengzhang, zhangxuejun, zhoujun, yanyonghong\}@hccl.ioa.ac.cn
}

\begin{document}

\maketitle 
\begin{abstract}
Incremental language learning with pseudo-data can alleviate catastrophic forgetting in neural networks. However, to obtain better performance, former methods have higher demands for pseudo-data of the previous tasks. The performance dramatically decreases when fewer pseudo-data are employed. In addition, the distribution of pseudo-data gradually deviates from the real data with the sequential learning of different tasks. The deviation will be greater with more tasks learned, which results in more serious catastrophic forgetting. To address these issues, we propose reminding incremental language model via \textbf{d}ata-\textbf{f}ree \textbf{s}elf-\textbf{d}istillation (\textbf{DFSD}), which includes self-distillation based on the Earth Mover’s Distance and hidden data augmentation. By estimating the knowledge distribution in all layers of GPT-2 and transforming it from teacher model to student model, the Self-distillation based on the Earth Mover’s Distance can significantly reduce the demand for pseudo-data. Hidden data augmentation can greatly alleviate the catastrophic forgetting caused by deviations via modeling the generation of pseudo-data as a hidden data augmentation process, where each sample is a mixture of all trained task data. The experimental results demonstrate that our DFSD can exceed the previous state-of-the-art methods even if the maximum decrease in pseudo-data is 90\%.
\end{abstract}

\section{Introduction}
The process of human learning is a long-term, continuous behavior. When learning new knowledge, humans will not obviously forget what they have learned. However, the traditional machine learning paradigm is unable to retain previously learned knowledge when learning new knowledge, which is referred to as catastrophic forgetting \citep{mccloskey1989catastrophic,french1999catastrophic}. Some people who are good at learning can learn new knowledge and consolidate old knowledge through analogy and memory. The goal of incremental learning is that the model can separately learn different tasks, without retrieving the data of previous tasks, and regardless of the order of tasks. At the end of learning the new task, the model can perform well in the new task and previously learned tasks. Since most training methods of neural networks are data-driven, a lot of works on mitigating  catastrophic forgetting are based on appropriately relaxing the restrictions on employing the data of previous tasks. In this paper, we investigate solving the catastrophic forgetting problem in incremental learning following data-free constraints.

Unifying the data format of different tasks is very beneficial to incremental learning. DecaNLP \citep{decaNLP} regards various NLP tasks as question answering (QA). Recently, LAMOL \citep{sun2019lamol} has been a generative lifelong learning framework, that uses a language model (LM) to learn various kinds of NLP tasks in QA-style, such as sentiment classification, goal-oriented dialogue systems, semantic role labeling, semantic parsing, and question answering. In LAMOL, the pseudo-data generated by the model is trained together with the new task, instead of the real data from the previous tasks.  L2KD \citep{l2kd} improves LAMOL by distilling the output of the model at the word-level and sequence-level. However, L2KD needs to train a single-task model as a teacher model when learning a new task. DnR \citep{dnr} improved LAMOL by distilling the attention and hidden states of some layers in GPT-2 \citep{gpt2}. Although these two methods can better alleviate catastrophic forgetting, they still need to generate pseudo-data equivalent to 20\% of the new task dataset size.

It is significant to solve the catastrophic forgetting problem in incremental learning following data-free constraints. In above-mentioned works, the performance dramatically decreases when fewer pseudo-data were trained together with a new task for the following two reasons: (1) the knowledge distribution of tasks in different layers of GPT-2 is not considered, which renders training of pseudo-data inefficient and causes obvious catastrophic forgetting when fewer pseudo-data. (2) with an increase in learned tasks and the imbalance of different task data, the deviation between the distribution of pseudo-data of the tasks and the distribution of real data of the tasks will gradually increase. Any deviation was regarded as ``chaos'' in \citep{sun2019lamol}. We define this deviation as noise, which will cause catastrophic forgetting.

To address these two issues, we propose a reminding incremental language model via data-free self-distillation (DFSD) which leverages self-distillation based on the Earth Mover’s Distance (EMD) \citep{emd} and hidden data enhancement (HDA). (1) The self-distillation based on EMD, regards the model trained on all previous tasks as the teacher (T) and the model training on a new task as the student (S) initialized by parameters of T. The student model needs to learn new tasks and retains the performance for old tasks. 

(2) \textbf{H}idden \textbf{D}ata \textbf{A}ugmentation (HDA) models the generation of pseudo-data as a hidden data augmentation process, where each sample is a mixture of all trained task data and is able to correct deviations.

DFSD is inspired by the knowledge distillation method in model compression  \citep{furlanello2018born,bert-emd} and noise correction \citep{arazo2019unsupervised}. In contrast to previous approaches, DFSD is a self-learning method that considers both the commonness and difference of knowledge distribution among different tasks in the model and fully utilizes noise in pseudo-data. Our incremental language model is both a teacher and a student. As our model is both a teacher and a student, we do not need to train a single-task model as a teacher.

Our main contributions in this paper are detailed as follows: 
(1) We propose DFSD, which can learn more useful knowledge from a small amount of noisy pseudo-data for incremental language learning and significantly outperforms previous state-of-the-art methods.
(2) We propose EMD-based self-distillation to estimate the distribution of task knowledge in all layers; it can significantly reduce the dependence on data replay. 
(3) We propose HDA to alleviate catastrophic forgetting caused by the deviation of the pseudo-data probability distribution from the real distribution.
(4) We analyzed the influence of HDA and self-distillation based on EMD on the final performance of DFSD. Experiments indicate that these process complement each other and can alleviate catastrophic forgetting from different aspects.

We will open-source our code after publishing the paper to facilitate further incremental language learning.

\section{Related Work}
Incremental learning is an essential step in promoting the realization of general artificial intelligence. Previous studies have explored how to alleviate catastrophic forgetting from different perspectives. EWC \citep{kirkpatrick2017ewc} and MAS \citep{aljundi2018mas} use regular methods to estimate the importance of parameters, and update the parameters based on these methods. Knowledge distillation has been researched in the field of incremental learning. Meta-MbPA \citep{meta-mbpa}, MbPA++ \citep{d2019mbpa}, GEM \citep{lopez2017gem} and A-GEM  \citep{chaudhry2018a-gem} use a small amount of data from old tasks, and use them to alleviate catastrophic forgetting. GEM and A-GEM uses these data to join the training phase, while MbPA++, in addition to training, will also be utilized in the inference phase. Meta-MbPA applies a meta-lifelong framework to improve MbPA++. In the field of computer vision, 
\begin{figure}[]
\centering
\includegraphics[width=\columnwidth]{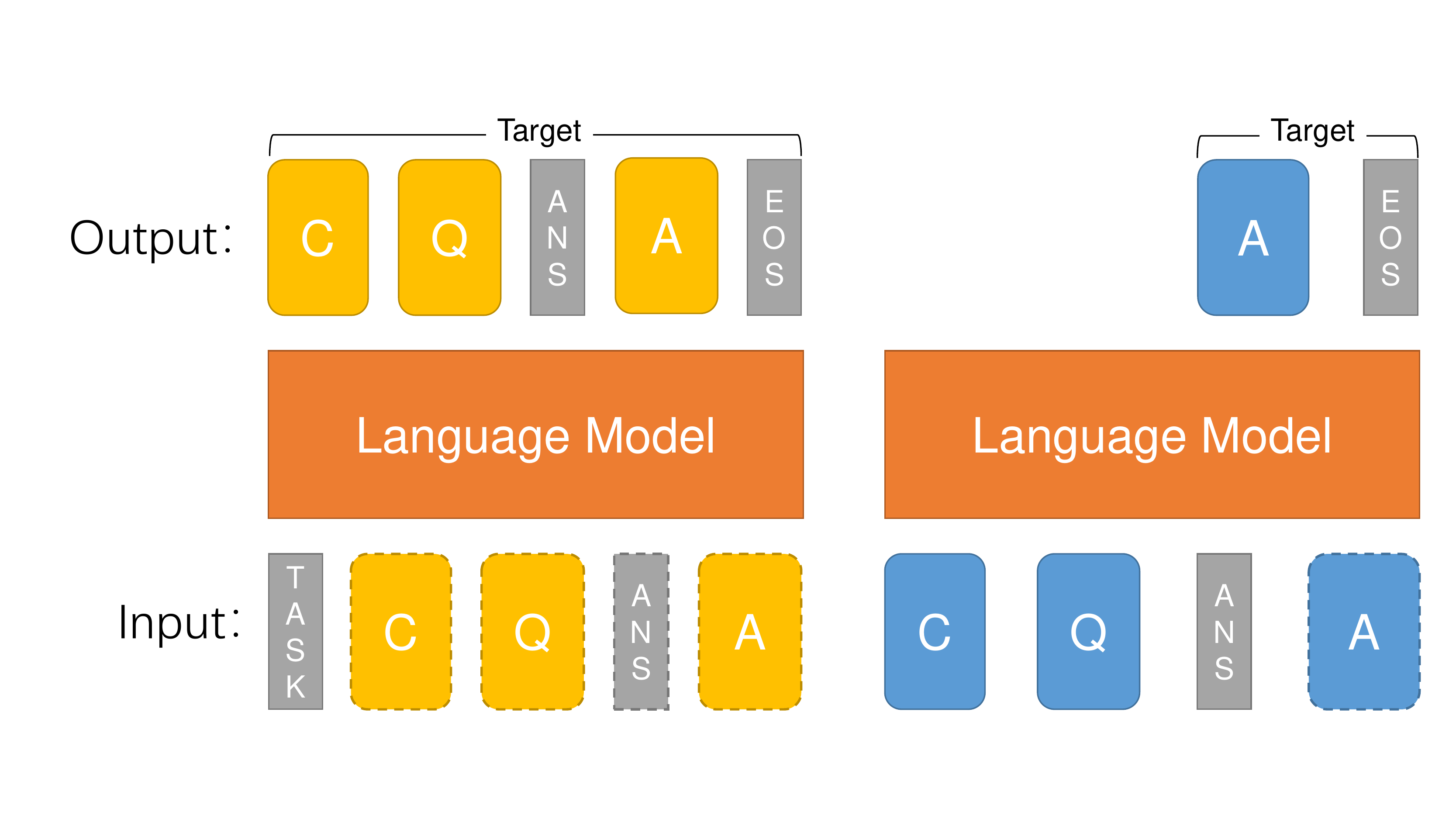} 
\vspace{-0.6cm}\caption{Illustration of two tasks in LAMOL. \textbf{Left}: LM task. \textbf{Right}: QA Task.}\vspace{-0.6cm}
\label{fig1}
\end{figure}
\citep{li2017lwf,van2018dgr+distill,zhai2019lifelong-gan} researched memory replay- or generation-based approaches with distillation. In the field of natural language processing, L2KD \citep{l2kd} uses word-level and sequence-level knowledge distillation to transform information from a teacher model to a student model which contains knowledge of previous tasks. L2KD needs to train a single-task model with a new task as a teacher model before incremental learning. DnR \citep{dnr} distills the attention and hidden state of some layers in GPT-2 \citep{gpt2}. Both L2KD and DnR use knowledge distillation to improve LAMOL \citep{sun2019lamol}. In contrast to prior work, we consider the distribution of tasks learned by each layer in GPT-2, which can make pseudo-data be learned more efficiently. Moreover, we are the first researchers to propose modeling generation of each pseudo-sample as a mixture of all learned tasks, and use it as a basis to correct the noise in the pseudo-data. Therefore, both methods are complementary to each other and can be applied to other fields.

\section{Methodology}
The proposed DFSD is based on LAMOL \citep{sun2019lamol}. Before we introduce our method in detail, we will briefly introduce it.

LAMOL uses a single language model to complete QA and LM tasks. The QA task refers to a sequence containing the context (C) and question (Q) as input, and the corresponding answer (A) is generated by the model. The LM task is to generate a complete sentence that contains C+Q+A, starting from a task-specific token. These two tasks are illustrated in Figure 1. Assume a order of tasks \{$D_{1}, D_{2}, ...$\}, where we do not know the number of tasks. When training on a new task $D_{\tau}, \tau>1$, the task-specific token will be applied as the first token to generate pseudo-samples that are trained with a new task. The number of pseudo-data is $\gamma|D_{\tau}|$, where $\gamma$ denotes the sampling ratio and $|D_{\tau}|$ is the number of data in task $D_{\tau}$. Since all $\tau-1$ tasks share the $\gamma|D_{\tau}|$ pseudo-samples, the model generates $\frac{1}{\tau-1}|D_{\tau}|$ pseudo-samples for the previous $\tau-1$ tasks.
\begin{figure}[ht] 
\centering 
\includegraphics[scale=0.2]{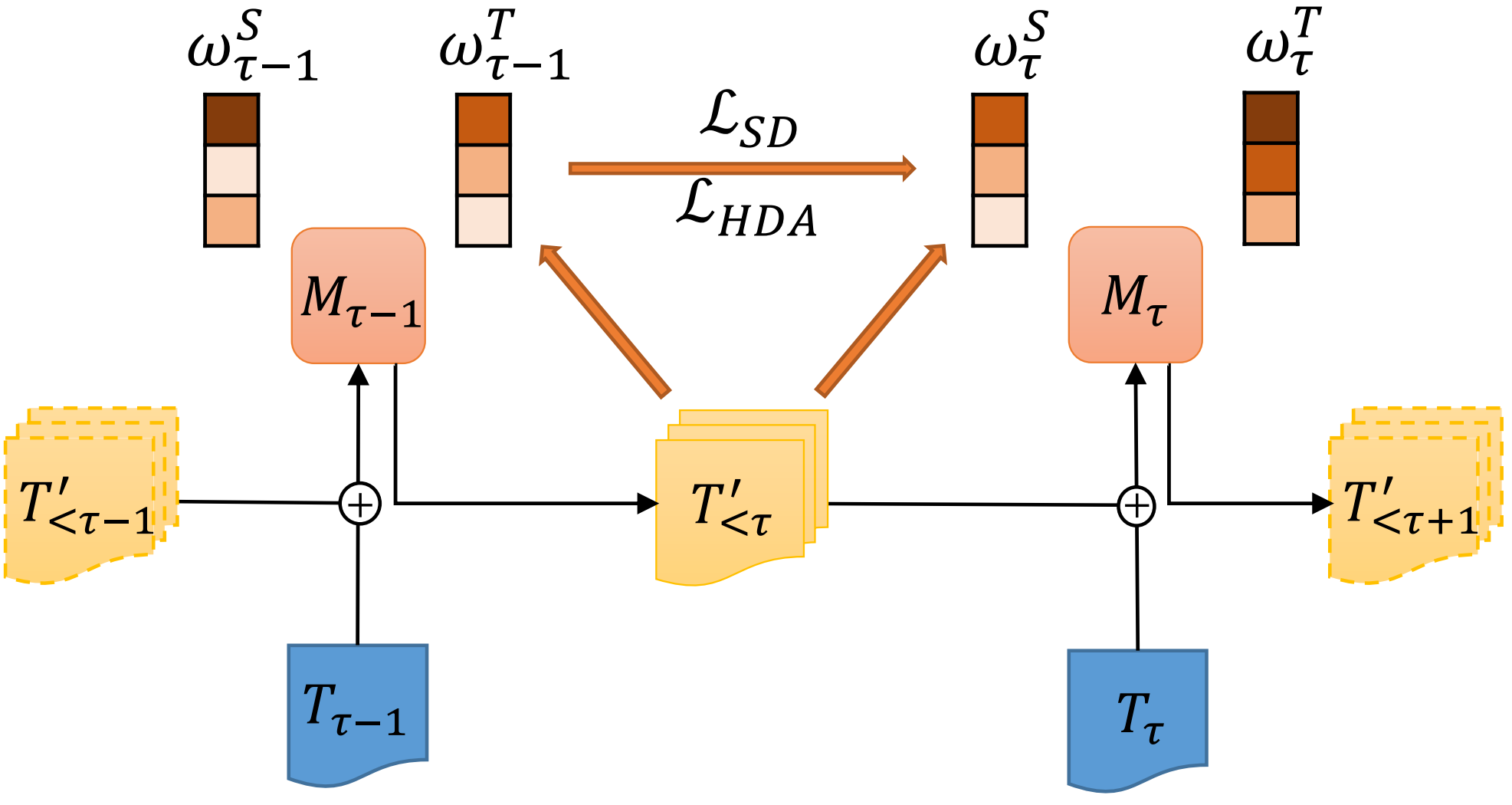} 
\caption{Framework of DFSD. $\omega$ indicates the distribution of task knowledge in all layers. $T$ and $S$ in $\omega$'s superscript denote the current model as the teacher or student, respectively, while subscripts denote the index of the task. $M_{\tau}$ indicates that the model has learned the $\tau$-th task. $T_{\tau}$ indicates the real data of the $\tau$-th task. $T_{<\tau}^{'}$ indicates the pseudo-data of previous $\tau-1$ tasks.} \vspace{-0.6cm}
\label{fig2} 
\end{figure}

\subsection{Overall Framework}
One of the main ideas in our proposed DFSD is to adaptively transfer different layers of knowledge from the teacher model to the student model when learning a new task since the knowledge learned in different tasks is distributed differently in different layers. The knowledge distribution of a task will shift when the model learns a new task. As we can not predict what the next task will look like during incremental learning, it will become very expensive to manually design the mapping relationship between the teacher layers and the student layers. It is significant to find a way to obtain the mapping relationship that can transform the knowledge from T to S with minimum cost. Moreover, from the aspect of data restriction, the incremental model needs to utilize the data of the previously learned tasks as little as possible when training a new task. The EMD \citep{emd} can measure the discrepancy between two distributions as transforming knowledge between two distributions with minimal accumulated cost. As the pseudo-data can be regarded as containing the distribution of tasks that have been learned, feeding a few pseudo-data into the model can obtain knowledge distribution of tasks. Therefore, we propose a method of self-distillation based on the EMD to transfer knowledge from the teacher model to the student model with pseudo-data. Both the teacher model and the student model share all parameters.

In addition, we innovatively propose applying hidden data augmentation (HDA) to solve the noise in pseudo-samples. There is various noise mixed in pseudo-samples, as generated by the model that has learned various kinds of tasks, degrading the performance of the model if we train them directly. This situation can be regarded as knowledge confusion similar to the situation in which humans have learned various branches of knowledge. Therefore, we model the generation of pseudo-data as a mixture of data of all learned tasks. 

These aforementioned methods are presented in Figure 2.

\subsection{Self-Distillation based on EMD}
In normal knowledge distillation, both the teacher model and the student model are trained with the same real dataset. However, it is not convenient to obtain these real datasets during incremental learning. Therefore, pseudo-data can be regarded as a sampling of the data distribution of the previous task.

Next, we conduct our self-distillation from three levels, namely, the embedding layer, attention, and the hidden state.

\textbf{Embedding Layer}  Word embedding, the most fundamental component among the various tasks of NLP, can map words into semantic space. 

The semantics of words are affected by context and change with different tasks. We minimize the semantic distance of words with similar contexts between T and S with the mean squared error.
Let $X=\left ( x_1,...,x_i,...,x_N\right )$ denote a pseudo-sample $X$ with length N; let $E^T=\left ( \mathit{e_1^T,...,e_i^T,...e_M^T} \right )$ denote the embedding of $X$ in the teacher model and let $E^S=\left ( \mathit{e_1^S,...,e_i^S,...e_M^S} \right )$ denote the embedding of $X$ in student model.
\begin{equation}
\mathbb{L}_{emb}= \frac{1}{M}\sum_{i}^{N} \left \|\mathit{e}_{i}^{T}-\mathit{e}_{i}^{S}  \right \|^2
\end{equation}
where $M$ means the dimension of a word vector.

\textbf{Attenion and Hidden State} GPT2 is stacked by multiple transformer decoders with a masked self-attention layer which makes the model pay attention only to the information before the current position. We transform linguistic knowledge, which is distributed in the masked attention matrix and hidden state matrix of different layers, from T to S based on EMD. Let $\mathbb{U^T}=\{(U_1^T,\omega_1^T),...,(U_K^T,\omega_K^T)\}$ denote the matrix and weights of all layers in the teacher model and $\mathbb{U^S}=\{(U_1^S,\omega_1^S),...,(U_K^S,\omega_K^S)\}$ denote the matrix and weights of all layers in the student model. Next, we let $\mathbb{D}=\{d_{ij}\}$ denote the cost of transforming attention or hidden state knowledge from $U_i^T$ to $U_j^S$. As Jensen–Shannon divergence(JSD) is symmetric and can be used to measure the difference between two different distributions, we apply it to calculate $d_{ij}$:
\begin{equation}
    d_{ij}=\frac{1}{2}\left ( U_{i}^{T}log\frac{U_{i}^{T}}{U_{j}^{S}} + U_{j}^{S}log\frac{U_{j}^{S}}{U_{i}^{T}} \right )
\end{equation}
where $i,j\in [1,K]$.

Next, we discover a way to minimize the cumulative transformation cost to complete the knowledge conversion from T to S by solving the following optimization problems:
\begin{eqnarray}
    &&min\sum_{i}^{K}\sum_{j}^{K}d_{ij}f_{ij}
\end{eqnarray}
where $f_{ij}$ means the mapping flow from the $i$-th layer of T to the $j$-th layer of S.  
Therefore, we define the result of EMD as follows:
\begin{eqnarray}
    \textbf{EMD}(U^T,U^S)=\frac{\sum_{i}^{K}\sum_{j}^{K}d_{ij}f_{ij}}{\sum_{i}^{K}\sum_{j}^{K}f_{ij}}
\end{eqnarray}
Both the attention matrix and the hidden state matrix can be calculated with \textbf{Eq.(4)}. The objective function consists of:
\begin{eqnarray}
\mathbb{L}_{SD}=\textbf{EMD}(\textbf{A}^T,\textbf{A}^S)+\textbf{EMD}(\textbf{H}^T,\textbf{H}^S)
\end{eqnarray}
where ${A}^T$ and ${A}^S$ are the attention matrices of T and S, respectively, and ${H}^T$ and ${H}^S$ are the hidden state matrices of T and S, respectively.

\begin{figure}[ht] 
\centering 
\includegraphics[scale=0.2]{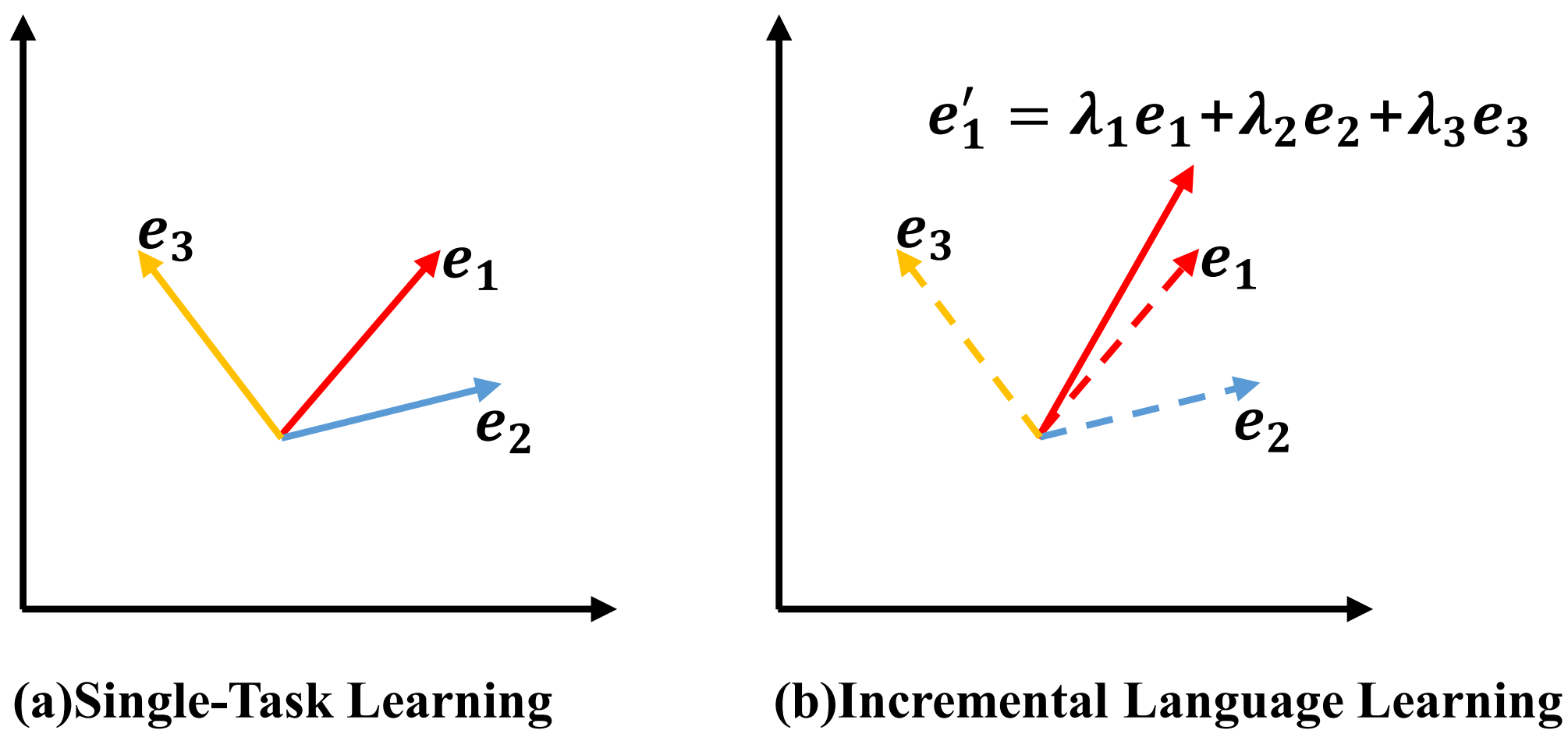} 
\caption{(a) Embedding of the task-specific tokens when each task is learned by single-task learning. (b) Embedding of the task-specific tokens when a stream of tasks is trained sequentially by incremental language learning.} \vspace{-0.6cm}
\label{fig3} 
\end{figure}

Without any prior knowledge of the task, we initialize $\omega_{j}^T$ and $\omega_{j}^S$ with $\frac{1}{K}$. However, the weights of different kinds of NLP tasks at different layers differ. Therefore, we follow the cost attention \citep{bert-emd} to update $\omega_{j}^T$ and $\omega_{j}^S$ to more accurately map the various linguistic knowledge of each layer.

\subsection{Hidden Data Augmentation}

We propose hidden data augmentation(HDA) to correct the problem that the distribution of task-specified pseudo-data is quite different from its corresponding real data distribution. Ideally, the task-specified token $\textbf{[TASK]}$, as the starting token to generate pseudo-samples, can produce task-specified pseudo-samples $\chi_{task}$ that are similar to task-specified real sample $\chi_{task}^{real}$ in content and grammar. However, previous task-specified token embedding will shift with the learning of new tasks, which results in uncontrollable scale noise in pseudo-samples, as illustrated in Figure 3 and visualized in Appendix. Since this noise is derived from the learning of new tasks, we regard the offset of $\textbf{[TASK]}$ in the semantic space as a data mixing process. 
Formally, let $\chi_{t<\tau}=\{\chi_{1}^{\tau},...,\chi_{\tau-1}^{\tau}\}$ denote all pseudo-samples of previous tasks, let $\Psi_{\tau}=\{\psi_{1},...,\psi_{\tau}\}$ denote the set of task-specified tokens, and let $\Phi_{\tau}=\{\phi_{1},...,\phi_{\tau}\}$ denote the distribution in the semantic space of tasks, where $\tau$ means the order of the current task. The generation of pseudo-data is expressed as follows:
\begin{eqnarray}
    &&\chi_{i}^{\tau}=P(\psi_i|\Phi_{\tau-1})
\end{eqnarray}
Eq. (6) means that the pseudo-sample generation of each task will be affected by the tasks that have been learned. Therefore, information from other tasks is mixed into $\chi_{i}^{\tau}$ in a special ratio before training the $\tau$-th task. When define $\Lambda_{i}^{\tau}=\{\lambda_1^{i},...,\lambda_{\tau-1}^{i}\}$ to represent the mixed weight of each previous-task data when generating pseudo-samples for the $i$-th task before training the $\tau$-th task. $\Lambda^{\tau}=\{\Lambda_{1}^{\tau},...,\Lambda_{\tau-1}^{\tau}\}$ is the set of all previous-task weight. Next, we regard $i$-th task-pseudo-sample generated as follow:
\begin{eqnarray}
    &&\hat{x}_{i}=\sum_{j=1}^{\tau-1}\lambda_j^{i}x_j^{i}
\end{eqnarray}
where $\Theta^T$ represents the parameters of T and $x_j^{i}$ represents the pseudo-sample generated by the model only training on $j$-th task. The generation of $x_j^{i}$ is an unobservable hidden process that we assume. Since this mixing process is automatically performed by the model, we cannot know the true value of $\Lambda^{\tau}$ so we refer to it as \textbf{HDA}. Therefore, each word in the pseudo-sample can be regarded as a mixture of previous-task information and can alleviate S forgetting the knowledge of the previous task. We consider these words pseudo-authentic labels provided by T to S. The objective function of HDA can be defined as follows:
\begin{equation}
    \mathbb{L}_{HDA}=-\sum_{i=1}^{\tau-1}\sum_{j=1}^{V}(\alpha\hat{h}_{ij}^{T}+(1-\alpha)\hat{x}_{ij})^{\textbf{T}}log(\hat{h}_{ij}^{S})
\end{equation}
where $\alpha$ denotes the confidence of T for the pseudo-samples as input, and $\hat{h}_{ij}^{T}$ and $\hat{h}_{ij}^{S}$ are the predicted temperature-softmax probabilities with T and S, respectively. $V$ is the vocabulary size of our model.

\subsection{Training Process}
 
As the QA and LM tasks are trained simultaneously in LAMOL, the methods mentioned in the previous two sections are also applied to the two tasks. Our complete objective function is expressed as follows:
\begin{equation}
    \begin{split}
        \mathbb{L}=& \mathbb{L}^{QA}+\beta\mathbb{L}^{LM}+\mu(\mathbb{L}_{SD}^{QA}+\mathbb{L}_{SD}^{LM})+\delta(\mathbb{L}_{HDA}^{QA}+\mathbb{L}_{HDA}^{LM})
    \end{split}
\end{equation}

where $\beta$ denotes the factor of the LM task, and $\delta$ and $\mu$ denote the factors of hidden data augmentation and self-distillation based on EMD, respectively. The training process is detailed in Appendix.
    
\section{Experiments Setup}

 To make the results comparable, we build our DFSD based on the implementation of LAMOL\footnote{\url{https://github.com/jojotenya/LAMOL}}. We use a task-specified token as the beginning token of pseudo-samples. The pseudo-samples ratio is $\gamma$ $\in$ [0.01,0.02,0.03.0.05]. $\delta$ and $\mu$, which are the factors of hidden data augmentation and self-distillation based on EMD, are 0.08 and 0.5 respectively. We set the learning rate to 1e-4 and set $\alpha$, the confidence of T in HDA, to 0.9.
\subsection{Experimental Data}
Following the setting of \citet{sun2019lamol}, we use five different NLP tasks in decaNLP \citep{decaNLP} to evaluate our proposed DFSD: SQuAD \citep{rajpurkar2016squad}, WikiSQL \citep{zhong2017wikisql}, WOZ \citep{wen2016woz}, QA-SRL \citep{he2017srl}, SST \citep{radford2017sst}. The five previously introduced datasets are five different NLP tasks that will be trained in random order to evaluate whether our proposed method generally comprises a variety of NLP tasks. In addition, Amazon, AGNews, DBPedia, Yahoo and Yelp are five classic text classification datasets\citep{DBLP:conf/nips/ZhangZL15}. We use a balanced version in \citep{d2019mbpa}. More detail is provided in Appendix.

\begin{figure}[ht] 
\centering 
\includegraphics[scale=0.48]{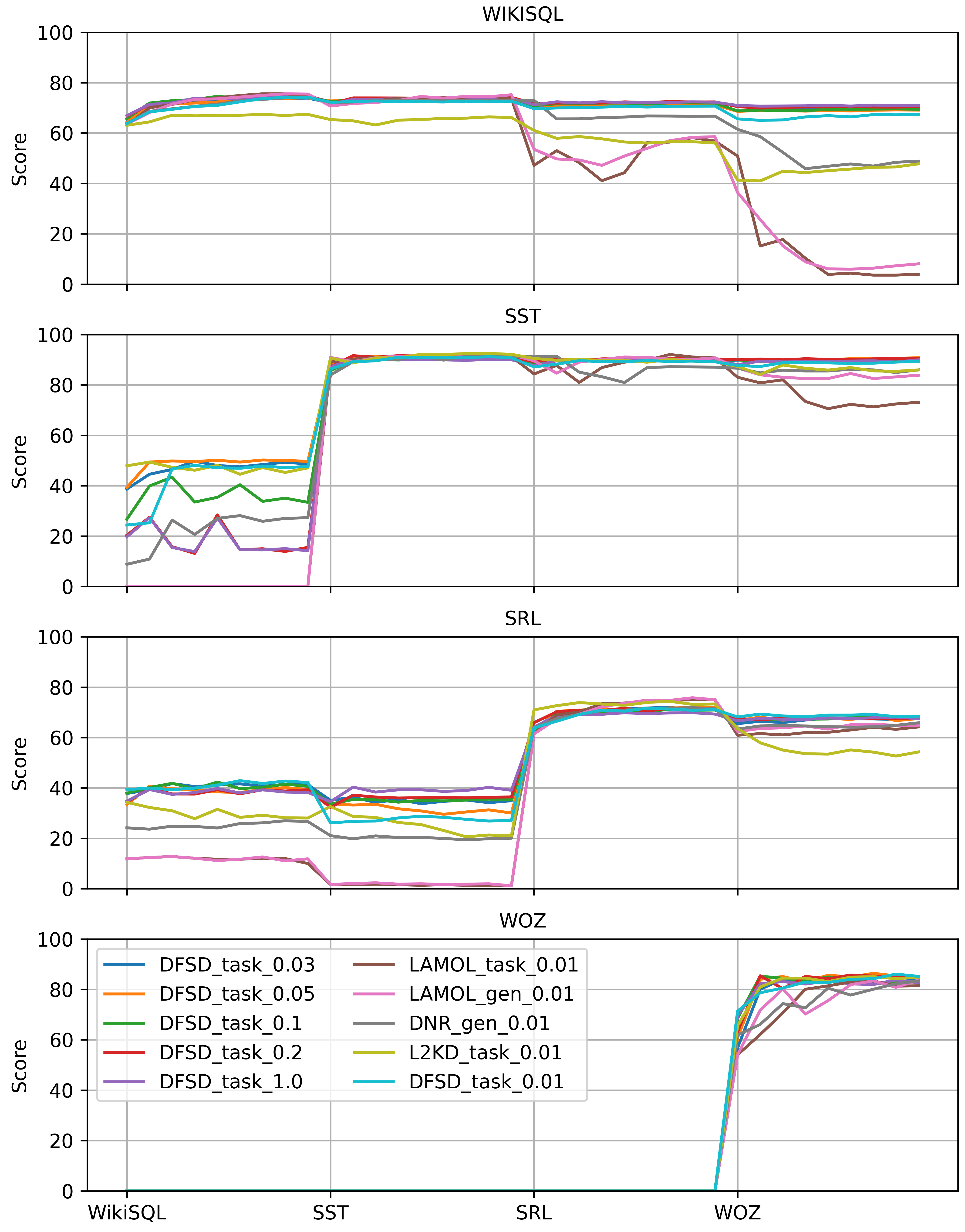} 
\caption{Models' performance after each epoch in the sequence SQuAD-WikiSQL-SST-SRL-WOZ. The format of the legend is "[model]\_[task-specified token]\_[sample ratio $\gamma$]".} \vspace{-0.6cm}
\label{deca5} 
\end{figure}

\subsection{Baselines}
\begin{itemize}
    \item \textbf{Fine-tuned}: Fine-tune GPT-2 one by one directly, according to the task order.
    \item \textbf{LAMOL}: Only QA and LM tasks are used to train the general incremental language model. We take task-specified token and [GEN] token of LAMOL to compare.
    \item \textbf{L2KD}: An improved version of LAMOL with normal distillation that trains a single-task model with the current task as the teacher model and a model trained on the previous task as the student model. It use the task-specified token.
    \item \textbf{DnR}: An improved version of LAMOL with distillation and replay that distills part of the layer. We take its best implementation of distilling the last two layers with naive matching. It use the [GEN] token.
    \item \textbf{MbPA++}: MbPA++ chooses parts of data from old tasks as episodic memory and replays it afterward. We choose its best setting and performance for text classification.
    \item \textbf{Meta-MbPA}: Meta-MbPA applies a meta-lifelong framework to improve MbPA++. 
\end{itemize}

\section{Results and Analysis}

\subsection{Three DecaNLP Tasks}
\begin{table*}[]
\centering
\begin{tabular}{@{}cccccccc@{}}
\toprule
\multirow{2}{*}{Methods} & \multicolumn{2}{c}{SST}                                   & \multicolumn{2}{c}{SRL}                                   & \multicolumn{2}{c}{WOZ}                                          & \multicolumn{1}{l}{\multirow{2}{*}{Avg.}} \\
                         & \multicolumn{1}{l}{SRL-WOZ} & \multicolumn{1}{l}{WOZ-SRL} & \multicolumn{1}{l}{SST-WOZ} & \multicolumn{1}{l}{WOZ-SST} & \multicolumn{1}{l}{SST-SRL} & \multicolumn{1}{l}{SRL-SST}        & \multicolumn{1}{l}{}                      \\ \midrule
Fine-tuned               & 50.2                        & 24.7                        & 62.9                        & 31.3                        & 32.8                        & \multicolumn{1}{c|}{33.9}          & 39.3                                      \\

LAMOL$_{0.05}^{TASK}$                & 77.3                        & 76.9                        & 78.1                        & 74.7                        & 73.4                        & \multicolumn{1}{c|}{75.8}          & 76.0                                      \\
LAMOL$_{0.2}^{TASK}$                  & 79.4                        & 79.9                        & 80.1                        & 78.7                        & 79.8                        & \multicolumn{1}{c|}{79.0}            & 79.5                                      \\
LAMOL$_{0.2}^{GEN}$                  & 80.0                        & 80.7                        & 79.6                        & 78.7                        & 78.4                        & \multicolumn{1}{c|}{80.5}            & 79.7                                      \\
DnR$_{0.2}^{GEN}$                      & 81.4                        & 82.1                        & 81.2               & 81.1                        & 80.9                        & \multicolumn{1}{c|}{81.6}          & 81.4                                      \\ \midrule
DFSD$_{0.01}$                     & 80.7                        & 79.9                        & 77.8                        & 77.4                        & 72.7                        & \multicolumn{1}{c|}{73.1}          & 76.9                                      \\
DFSD$_{0.02}$                     & 81.3                        & 80.8                        & 79.6                        & 78.9                        & 79.3                        & \multicolumn{1}{c|}{78.4}          & 79.7                                      \\
DFSD$_{0.03}$                     & 81.2                        & 81.4                        & 80.2                        & 80.1                        & 79.9                        & \multicolumn{1}{c|}{80.4}          & 80.5                                      \\
DFSD$_{0.05}$                     & \textbf{81.5}               & \textbf{82.4}                        & 80.8                        & 80.5                        & \textbf{81.7}                        & \multicolumn{1}{c|}{81.5}          & \underline{\textbf{81.4}}                                      \\ \midrule

MTL                      & \multicolumn{7}{c}{81.5}                                                                                                                                                                                                             \\ \bottomrule
\end{tabular}
\caption{Each column represents a permutation of [SST, SRL, and WOZ]. After training in this order, the model of the last task is averaged on the three tasks, and the subscript represents the value of $\gamma$. Avg. represents the average value of all training sequences. MTL represents the multi-task learning performance. The experimental results of the baseline are retrieved from \citep{sun2019lamol} and \citep{dnr}.}\vspace{-0.6cm}
\end{table*}
To comprehensively compare the performance of our proposed method, we firstly select three data sets [SST, SRL, and WOZ] in decaNLP. To eliminate the influence of the training order, we train them in six orders. When training each order, we test the model after the last task has been trained and take the average of the three task scores as the performance. The average of all orders is taken as the performance of the current model. All the results are shown in Table 1.

As shown in Table 1, we determine that the fine-tuned method performs poorly on these three task data sets, and the order of learning has a greater impact on the performance of the task. As a baseline of the incremental language model, we can see that LAMOL has a better performance when $\gamma=0.2$, which is only 2-3\% lower than MTL; however, when $\gamma=0.05$, it is quite different from MTL, reaching 7\%. According to \citep{sun2019lamol}, when the $\gamma$ is lower, the performance of the model will decrease significantly. When the $\gamma$ approaches 0, its performance is similar to that of fine-tuning. Compared with LAMOL, DnR has a certain improvement in each sequence, and its overall performance is almost equivalent that of MTL, which indicates that it is effective for distilling only the last two layers. We determine that our proposed method is better than LAMOL under the same gamma. When $\gamma=0.01$, DFSD$_{0.01}$ still achieves satisfactory performance, and it is even better than LAMOL$_{TASK}^{0.05}$ when $\gamma$ decreased by 80\%. When $\gamma=0.02$, DFSD$_{0.02}$ achieves the same performance as LAMOL$_{GEN}^{0.2}$ with a reduction in $\gamma$ by 90\%, indicating that our method can still accurately capture information of the previous tasks with lower-resource pseudo-data.

With an increase in $\gamma$, the performance of our method gradually improves. When it is increased to $\gamma=0.05$, DFSD$_{0.05}$ can achieve the performance of DnR with with a reduction in $\gamma$ by 60\%. We believe that this outcome occurs because DFSD fully considers the semantic distribution of different tasks in different layers when distilling and allows the model to automatically transfer knowledge based on this distribution, which produces more efficient training of pseudo-data. This issue is analyzed in latter visualization. 

\begin{table}[htp]
\centering 
\begin{tabular}{@{}ccccccc@{}}
\toprule
Sub-order       & 234         & 243         & 324             & 342          & 423          & 432                      \\ \midrule
01\#\#\#       & 74.8          & 75.4      & 75.1           & 75.2         & 75.1          & 75.6                     \\
10\#\#\#      & 75.7           & 75.5       & 75.2         & 75.1          & 75.2          & 75.1                     \\
\#\#\#01    & 75.0              & 75.2      & 75.4          & 75.1          & 75.1          & 75.1 \\
\#\#\#10     & 76.8           & 76.7        & 76.7          & 76.7       & 76.2           & 75.8                     \\ \bottomrule
\end{tabular}
\caption{Performance of all orders of [SQuAD (0), WikiSQL (1), SST (2), SRL (3), WOZ (4)] when SQuAD and WikiSQL are heads or tails. We train the model with these orders by our DFSD when $\gamma=0.05$. Each score in the table indicates that the model of the last task is averaged on the five tasks.}\vspace{-0.6cm}
\end{table}.

\subsection{Five DecaNLP Tasks}

The larger the number of learning tasks is, the earlier the learning tasks are more likely to be forgotten. We verify the effectiveness of DFSD by learning more tasks. We choose [SQuAD, WikiSQL, SST, SRL, and WOZ]. When SQuAD or WikiSQL learns as a nonfirst task, since both of them are significantly larger than others, the large amount of pseudo-data generated will make the model similar to multitask learning (the only difference is obtained with real data or pseudo-data). Therefore, the order with SQuAD and WikiSQL as the head will be more difficult to learn. We explore SQuAD and WikiSQL as the beginning or end in all orders; it contains 24 orders. As shown in Table 2, the experimental results are consistent with our expectations. According to the size of the dataset, the order from large to small is more challenging than the order from small to large. However, our method has robust performance for these different sequences. 

To indicate the effectiveness of our DFSD, we choose another smaller $\gamma$ and the most difficult order to compare with previous methods. The order, SQuAD-WikiSQL-SST-SRL-WOZ, is from the largest to the smallest. As $\gamma|D_{\tau}|$ decreases and pseudo-data need to be generated for more tasks, the order is more difficult to learn. As shown in Figure 4, our DFSD is also robust even if $\gamma=0.01$. When $\gamma=0.01$, we observe that previous methods forget WikiSQL as more tasks are learned, while DFSD is still effective, especially when learning WOZ. When learning WOZ, $\gamma|D_{WOZ}|=25$, there are only 6 pseudo-samples corresponding to each previous task. DFSD can efficiently learn and migrate knowledge from fewer pseudo-samples. DFSD is beneficial for the model to learn more tasks. In Figure 4, we also show the results of other higher $\gamma$ of DFSD. We can find that DFSD is highly robust to the $\gamma$ and can stably alleviate catastrophic forgetting. Moreover, it can be observed from the SST results that DFSD can migrate more knowledge from SQuAD and WikiSQL than other methods. The results of SRL show that DFSD can transfer more knowledge in SQuAD, WikiSQL and SST compared with other methods.

\subsection{Text Classification}
To show the effectiveness of our DFSD in training different datasets of one NLP task. We compare DFSD against with LAMOL, DnR, L2KD, MbPA++ and Meta-MbPA. For DFSD, when we train a new task, we only randomly sample 10\% of the data in each epoch due to a lack of computing resources. This means that although the model has been studied for 9 epochs, it is equivalent to learning the whole data set only once. Meanwhile, the same $\gamma$ is equivalent to 10\% of other methods in DFSD. We compared with $\gamma=0.2$ due to its better performance in LAMOL, L2KD and DnR. 
\begin{table}[htp]
\centering
\begin{tabular}{@{}llllll@{}}
\toprule
Order & i & ii & iii & iv & Avg. \\ \midrule
MbPA++$^{1.0}$ & 70.8 & 70.9 & 70.2 & 70.7 & 70.7 \\
Meta-MbPA$^{1.0}$ & 77.9 & 76.7 & 77.3 & 77.6 & 77.4 \\
LAMOL$_{0.2}^{1.0}$ & 76.7 & 77.2 & 76.1 & 76.1 & 76.5 \\
DnR$_{0.2}^{1.0}$ & 77.4 & 77.2 & 77.1 & 76.9 & 77.2 \\ \midrule
DFSD $_{0.2}^{0.1}$& \textbf{77.6} & \textbf{77.4} & \textbf{77.8} & \textbf{77.9} & \textbf{77.7} \\ \bottomrule
\end{tabular}
\caption{Summary of results on text classification. The superscript indicates the percent of the dataset used during training, and the subscript indicates the sampling ratio $\gamma$.}\vspace{-0.6cm}
\end{table}
As shown in Table 3, we discover that our DFSD can outperform previous state-of-arts methods. Moreover, we discover that our DFSD can make the old task better with the learning of the new task when training different datasets of one NLP task. We show the performance curve of the above five text classification tasks in Appendix.

\begin{figure}[ht] 
\centering 
\includegraphics[scale=0.185]{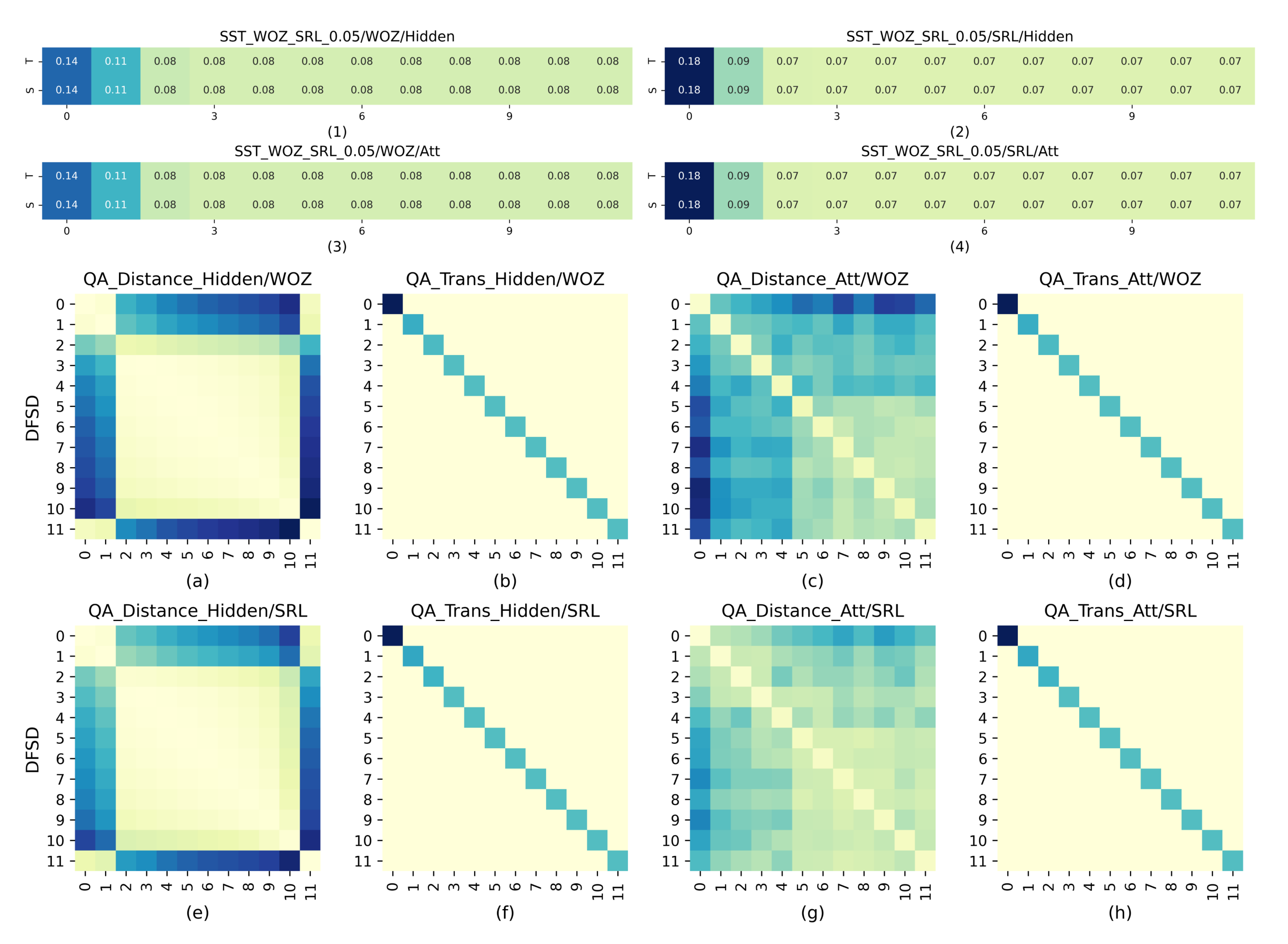} 
\caption{In the learning order of SST-WOZ-SRL, Figure 5(1)-(4) are the distribution of transferable knowledge of attention and the hidden states for each task in DFSD$_{0.05}$. Figure 5(a)(e) and Figure 5(b)(f) represent the distance of \textbf{H} of each layer between the teacher model and the student model, and the amount of knowledge transfer among all layers in the QA task. Figure 5(c)(g) and Figure 5(d)(h) illustrate the results of \textbf{A}.} \vspace{-0.6cm}
\label{trans} 
\end{figure}

\subsection{Visualization of Data-Free Self Distillation}

In the training process of self-distillation, we apply the EMD to obtain the knowledge mapping relationship between the layers of the teacher model and the layers of the student model because we believe that the knowledge distributed in each layer in GPT-2 also differs for different tasks. In Figure 5(1)-(4), we show the distribution of transferable knowledge of attention (\textbf{A}) and hidden states (\textbf{H}) in DFSD$_{0.05}$. We discover that the transferable knowledge weights of different tasks are mainly distributed in the first three layers, regardless of the order of the tasks. Note that the knowledge weight of the first layer is the highest in all orders. Our analysis indicates that this result is attributed to the notion that the knowledge extracted from the shallow layers has better generalization.  
It can be understood that the knowledge extracted from the shallow layers is shared by different tasks and that what the deeper layers learn is task-specific knowledge. The transferable knowledge distribution of all orders is almost consistent with Figure 5(1)-(4) and is provided in Appendix.

Regardless of whether \textbf{A} or \textbf{H}, we determine that the results of knowledge flow shown in Figure 5(b)(d)(f)(h) fall between the teacher layers and the corresponding student layers and that there is no cross-flow between different layers. This finding shows that the knowledge learned at each layer of GPT-2 is at different levels. In Figure 5(a)(e), which shows the distance distribution of \textbf{H} between the teacher model and the student model, we can observe that the first two layers and the last layer are very similar and the middle layers are similar, which proves that the features of the knowledge extracted by the shallow layers are similar. As the output layer needs to map the features to a vector of the length of the vocabulary, the knowledge contained in it is similar to that in the shallow layers. This process resembles a human learning process, in which, first, simple and popular knowledge is learned; second, complex knowledge is deeply learned; and last, complex knowledge is simplified and rendered popular. In Figure 5(c)(g), we can determine that the attentions of different layers are far apart, which indicates that attentions of different layers pay attention to the semantics of different levels.

\begin{table}[htp]
\centering
\begin{tabular}{@{}c|cccccc@{}}
\toprule
\multicolumn{1}{l|}{METH / $\gamma$} &0.001 &0.005     & 0.01          & 0.02          & 0.03          & 0.05 \\ \midrule
LAMOL$_{G}$  &52.0  &63.9     & 69.3          & 72.0           & 70.2          & 78.9$^*$ \\
LAMOL$_{T}$ &56.3 &63.9   & 63.3           & 68.0           & 75.5           & 76.9$^*$\\
LAMOL$_{R}$ &50.6 &64.5  & 65.8           & 73.0           & 74.9           & 78.9$^*$\\
DnR$_{NM}^{-2}$ &61.0 &68.2      & 71.9          & 78.4          & 78.0           & 78.1\\
L2KD &56.1 &66.5           & 72.3           & 70.8           & 76.8           & 77.4          \\ \midrule
DFSD &\textbf{66.8} &\textbf{77.9}                  & \textbf{79.9} & \textbf{80.8} & \textbf{81.4} & \textbf{82.4} \\
\multicolumn{1}{r|}{w/o SD} &54.5 &71.6  & 71.4          & 74.6          & \underline{79.1}    & \underline{79.0}    \\
\multicolumn{1}{r|}{w/o HDA} &\underline{62.0} &\underline{76.2} & \underline{77.9}    & \underline{78.6}    & \underline{79.4}    & \underline{79.5}\\ \bottomrule
\end{tabular}
\caption{The results of the ablation study when the order is SST-WOZ-SRL. The subscripts $T$ and $G$ indicate with task-specified and without task-specified token, respectively. $R$ indicates that real data are employed. * indicates that the experimental results are retrieved from \citep{sun2019lamol,l2kd,dnr}. We implement the remaining scores following corresponding paper.}\vspace{-0.6cm}
\end{table}

\subsection{Ablation Study}
To verify the respective effects of the two methods that we proposed, we conducted an ablation study and summarized the results in Table 4. When $\gamma \in$ [0.001, 0.005, 0.01, 0.02, 0.03, and 0.05], some experimental results are not in \citep{sun2019lamol,l2kd,dnr}; we reproduce them according to the corresponding paper.

We discover that with the decrease in $\gamma$, the catastrophic forgetting of LAMOL becomes increasingly serious regardless of real or pseudo-data are utilized. DnR and L2KD achieve some improvement over LAMOL, but there is still a significant performance degradation when $\gamma=0.01$. Both SD and HDA can obviously alleviate catastrophic forgetting, and the smaller the $\gamma$ is, the more significant it is. Especially when $\gamma\leq0.01$, SD can greatly alleviate catastrophic forgetting, while the performance of other methods has dropped sharply. Compared with LAMOL, SD can increase by a maximum of 14.6 percentage points, which indicates that SD can estimate knowledge distribution and complete the transfer of knowledge in GPT-2 through a very small amount of data. When only SD is utilized, it is better than other methods. This result indicates that SD has a higher utilization rate of knowledge than other methods. when $\gamma\geq0.005$, HDA is better than LAMOL with three different settings, indicating that HDA alleviates the catastrophic forgetting caused by the noise in the pseudo data. When $\gamma\leq0.02$, HAD is not as good as SD, but when $\gamma > 0.02$, it is better than SD. Our analysis is because when $\gamma$ is small, the positive impact of SD's knowledge transfer is greater than the negative impact of noise in the pseudo data. With the increase of $\gamma$, the influence of noise becomes more prominent, the positive influence of SD gradually reaches saturation, and the role of HDA is gradually manifested. This shows that when there are fewer pseudo data, lack of data and inefficient learning are the main factors of catastrophic forgetting. However, when there are sufficient pseudo data, the impact of data deviation on catastrophic forgetting will be more prominent. Therefore, SD and HDA are complementary to alleviate catastrophic forgetting.

\section{Conclusion}
In this paper, we propose data-free self-distillation (DFSD), which is a simple yet efficient method that achieves the same performance as previous state-of-the-art methods with a reduction in pseudo-data by 75-90\%.
We apply the EMD-based multilayer to multilayer self-distillation method to estimate the knowledge distribution of tasks in all layers and transfer it from the teacher model to the student model. The Experimental results show that the knowledge distribution of different tasks is concentrated in the first three layers of GPT-2 with 12 layers.
HDA can alleviate catastrophic forgetting caused by the deviation in the pseudo-data probability distribution from the real distribution. Both of them complement each other and have the potential to be applied to practical scenes.

\bibliography{anthology,custom}
\end{document}